\documentclass[letterpaper, 10 pt, conference]{ieeeconf}  

\IEEEoverridecommandlockouts                              

\usepackage{graphics} 
\usepackage{epsfig} 
\usepackage{mathptmx} 
\usepackage{times} 
\usepackage{amsmath} 
\usepackage{amssymb}  

\usepackage{algorithm}
\usepackage{algpseudocode}
\usepackage{tabularx}
\usepackage{float}     
\usepackage{placeins}
\usepackage{multirow}
\usepackage{booktabs}

\title{\LARGE \bf
BiTAA: A Bi-Task Adversarial Attack for Object Detection and Depth Estimation via 3D Gaussian Splatting}

\author{
Yixun Zhang, Feng Zhou and Jianqin Yin*\thanks{*Corresponding author.} \\
School of Intelligent Engineering and Automation,
Beijing University of Posts and Telecommunications, China \\
{\tt\small \{zhangyixun, zhoufeng, jqyin\}@bupt.edu.cn}
}

\begin{document}

\maketitle
\thispagestyle{empty}
\pagestyle{empty}

\begin{abstract}

Camera-based perception is critical to autonomous driving yet remains vulnerable to task-specific adversarial manipulations in object detection and monocular depth estimation. 
Most existing 2D/3D attacks are developed in task silos, lack mechanisms to induce controllable depth bias, and offer no standardized protocol to quantify cross-task transfer, leaving the interaction between detection and depth underexplored.
We present BiTAA, a bi-task adversarial attack built on 3D Gaussian Splatting that yields a single perturbation capable of simultaneously degrading detection and biasing monocular depth. Specifically, we introduce a dual-model attack framework that supports both full-image and patch settings and is compatible with common detectors and depth estimators, with optional expectation-over-transformation (EOT) for physical reality. 
In addition, we design a composite loss that couples detection suppression with a signed, magnitude-controlled log-depth bias within regions of interest (ROIs) enabling controllable near or far misperception while maintaining stable optimization across tasks. 
We also propose a unified evaluation protocol with cross-task transfer metrics and real-world evaluations, showing consistent cross-task degradation and a clear asymmetry between $\mathrm{Det}\!\rightarrow\!\mathrm{Depth}$ and from $\mathrm{Depth}\!\rightarrow\!\mathrm{Det}$ transfer. 
The results highlight practical risks for multi-task camera-only perception and motivate cross-task-aware defenses in autonomous driving scenarios.

\end{abstract}
    
\section{Introduction}
\label{sec:intro}

Modern autonomous driving systems~\cite{autoware,baidu,tesla} rely on camera-based perception, where object detection and monocular depth form the front end for downstream tracking, forecasting, and planning. Cameras are attractive for their cost and availability, and the two tasks largely share visual evidence such as textures, edges, and occlusions \cite{perceptionsurvey,monodepthsurvey}. As these modules feed into the decision chain, even moderate perception bias can accumulate and compromise safety \cite{attacksurvey}. This motivates studying adversarial robustness specifically for camera-based detection and depth, and understanding how perturbations propagate across tasks.

Recent adversarial studies on camera-based perception have advanced both digital and physical attacks on detectors, including full-image and patch formulations, white-/black-box regimes, and the use of expectation-over-transformation (EOT) to improve physical realism and reproducibility \cite{det_attack_survey,det_attack_eot,det_attack_phys}. For monocular depth, attacks typically inflate reconstruction errors or distort global scale, with progress from model-specific to model-agnostic settings \cite{depth_attack_survey,depth_attack_scale}. Beyond 2D pixels, differentiable-rendering approaches optimize parametric 3D representations (meshes, NeRFs, 3D Gaussians) and bridge to image-space losses, offering cross-view consistency and natural integration with physical constraints \cite{dr_mesh_attack,dr_nerf_attack,3dgs_original}. These lines of work provide strong baselines and practical toolkits, however, research is still largely organized in task silos, and cross-task effects between detection and depth remain under-studied \cite{cross_task_gap}.

To be more specific, detection and depth often share the same camera input and exploit overlapping visual cues. Yet most prior methods optimize for one task at a time, leaving open how to construct a single perturbation that, after multi-view rendering, consistently suppresses detection while also biasing monocular depth. The challenge is to coordinate potentially competing objectives within one parameter space and maintain stable gradients across views (\textbf{Challenge 1}). Besides, system-level risks frequently stem from consistent near/far misperception rather than indiscriminate error inflation. Existing depth attacks rarely provide signed and magnitude-controlled bias constrained to semantic regions of interest (ROIs). A practical design must balance physical realizability (e.g., color budgets) with geometric plausibility (e.g., shape stability) to avoid degenerate solutions (\textbf{Challenge 2}). Beyond per-task performance, practitioners need to know whether a perturbation optimized against cross-task transfer and whether the two directions differ. We denote this by $\mathrm{Det}\!\rightarrow\!\mathrm{Depth}$ as train on detection, evaluate on depth, and $\mathrm{Depth}\!\rightarrow\!\mathrm{Det}$ as train on depth, evaluate on detection. This motivates a standardized, model-agnostic protocol to quantify both directions under a fixed view set and consistent reporting statistics (\textbf{Challenge 3}).

To address these aforementioned challenges, we propose a novel bi-task adversarial attack built on 3D Gaussian Splatting (BiTAA) aimed at simultaneously achieving adversarial effectiveness in both detection and depth estimation tasks. First, we formulate a unified framework that directly optimizes a single perturbation in the 3D Gaussian parameter space and renders it across views via a fixed differentiable renderer. The task models are kept frozen for compatibility with common detectors and depth estimators, and EOT is optionally applied to the detection branch to model physical variability \cite{det_attack_eot}. All 14 degrees of freedom per Gaussian are updated, and view aggregation promotes cross-view consistency and stable gradients during joint optimization. This turns shared-input coupling into a single, end-to-end optimization problem in a geometry-aware parameterization, addressing \textbf{Challenge 1}. Second, we introduce a composite loss that couples detection suppression with a signed, magnitude-controlled log-depth bias restricted to per-view ROIs. The direction and amplitude are governed by a sign variable and a target magnitude; ROIs are constructed from detector boxes with inward shrinkage, and clean depth predictions are cached for stable residuals. To ensure physical plausibility and geometric coherence, we add shape-stability penalties on positions, scales and quaternions, together with realizability regularization via a color-space $L_{\infty}$ budget, total variation on residual images, and an optional high-frequency penalty. This design enables controllable near/far misperception while preserving plausible geometry and stable training dynamics, tackling \textbf{Challenge 2}. Third, we establish a unified cross-task transfer protocol comprising Det$\rightarrow$Depth, Depth$\rightarrow$Det, and Joint settings. The evaluation reports per-view maximum confidence for detection, ROI-averaged log-depth shifts with sign agreement for depth, and a normalized transfer-efficiency metric. The protocol uses a fixed view set and consistent statistics, facilitating reproducible comparisons across models and revealing directional asymmetry that informs cross-task-aware defenses, handling \textbf{Challenge 3}.

Our main contributions are as follows: 
\begin{itemize}
    \item We introduce BiTAA, a unified bi-task adversarial framework in a 3D Gaussian parameter space that jointly degrades detection and biases monocular depth after multi-view rendering, with frozen task models and optional EOT for physical variability. 
    \item We propose a composite loss that combines detection suppression with signed, magnitude-controlled log-depth bias within ROIs, together with shape-stability and realizability regularization for physical plausibility and numerical stability. 
    \item We propose a standardized cross-task transfer protocol with model-agnostic metrics for confidence suppression, ROI-averaged depth shifts, and normalized transfer efficiency, revealing clear directional asymmetry. The results of the experiment present the BiTAA's superior adversarial effectiveness both in detection (15.03\% mAP) and depth estimation (0.1989 AbsRel) task.
\end{itemize}

\section{Related Work}
\label{sec:related work}

\textbf{Camera-only perception for autonomous driving.}
Modern autonomous driving stacks rely on camera-based object detection and monocular depth to support downstream tracking, forecasting, and planning, where these modules share visual evidence such as textures, edges, and occlusions \cite{perceptionsurvey,monodepthsurvey}. Cameras remain attractive for their cost and availability, and the robustness of detection-depth perception is therefore central to overall system safety \cite{attacksurvey}.

\textbf{Adversarial attacks on detection and depth (2D).}
For camera-based detection, prior work~\cite{CAMOU,UPC,DAS,FCA} spans digital and physical settings, full-image and patch formulations, and white-/black-box regimes; expectation-over-transformation (EOT) is widely adopted to improve physical reality and evaluation reproducibility \cite{det_attack_eot,det_attack_patch,det_attack_phys}. These methods have demonstrated strong effectiveness on standard detectors and provide practical toolkits for evaluating robustness \cite{det_attack_survey}. For monocular depth, attacks typically increase reconstruction error or distort global scale, with steady progress from model-specific to model-agnostic formulations \cite{depth_attack_survey,depth_attack_scale}. However, ROI-constrained, signed and magnitude-controlled log-depth bias remains under-explored, and cross-task transfer is rarely quantified \cite{cross_task_gap}.

\textbf{3D differentiable-rendering based attacks.}
Beyond 2D pixels, recent approaches~\cite{TT3D,dr_mesh_attack,dr_nerf_attack,dr_3dgs_attack} optimize parametric 3D scene/object representations (meshes, NeRFs, and 3D Gaussians) and bridge to image-space losses via differentiable rendering. This paradigm offers cross-view consistency, integrates naturally with physical constraints and printing pipelines, and has shown promising lab-to-field behavior \cite{phys_consistency_survey}. Among these, 3D Gaussian splatting (3DGS) provides an efficient, gradient-friendly space for parameter updates \cite{3dgs_original}. Most existing 3D attacks, however, primarily target detection, with fewer studies treating detection and monocular depth jointly \cite{dr_3d_joint_gap}.

\textbf{Limitations of prior work.}
While the above lines of research have established powerful baselines and practical methodologies, there is still no unified setting where a single perturbation simultaneously suppresses detection and induces a signed, magnitude-controlled log-depth bias within semantic regions, nor a standardized protocol to measure cross-task transfer between detection and depth \cite{cross_task_protocol_gap}. Our study builds on these advances by formulating a bi-task attack in a 3D Gaussian parameter space and adopting a unified transfer protocol to systematically assess such interactions.

\section{Method}
\label{sec:method}

\subsection{Preliminaries}
\label{sec:prelim}

\textbf{3D Gaussian Splatting as a Backbone.}
We model the target object as a set of $N$ anisotropic 3D Gaussians
\begin{equation}
\mathcal{G}=\big\{(\mu_i,\ \alpha_i,\ s_i,\ q_i,\ c_i)\big\}_{i=1}^{N},
\end{equation}
where $\mu_i\!\in\!\mathbb{R}^3$ is the position, $\alpha_i\!\in\!\mathbb{R}_{+}$ the opacity, $s_i\!\in\!\mathbb{R}^3$ the axis-aligned scale, $q_i\!\in\!\mathbb{R}^4$ a unit quaternion (rotation), and $c_i\!\in\![0,1]^3$ the RGB color. Each Gaussian thus carries 14 degrees of freedom (DoFs). We treat 3DGS generation as a frozen backbone that, given a few posed images, provides an initial set $\mathcal{G}_0$.

\textbf{Differentiable Rendering \& Optimization.}
Given a calibrated view $v$, a fixed differentiable renderer $\mathcal{R}$ produces RGB and alpha from $\mathcal{G}$; we render a finite view set $V$ to promote multi-view consistency:

\begin{equation}
I_v=\mathcal{R}(\mathcal{G},v), \quad v\in V=\{v_1,\ldots,v_m\}. 
\end{equation}

Starting from $\mathcal{G}_0$, we optimize a (possibly masked) subset of the 14D parameters per Gaussian while keeping $\mathcal{R}$ fixed; image-space losses (introduced next) are aggregated over $V$ and back-propagated to $\mathcal{G}$:
\begin{equation}
\mathcal{G}=\mathcal{G}_0+\Delta\mathcal{G}, \qquad \Delta\mathcal{G}\in\mathbb{R}^{N\times 14}.
\end{equation}

Unless otherwise stated, all task networks used later are frozen; gradients flow from image-space losses through $\mathcal{R}$ to the parameters in $\mathcal{G}$. The concrete bi-task objectives and their aggregation across views $V$ will be introduced in the next sections.

\begin{figure*}[ht]
    \centering
    \includegraphics[width=\linewidth]{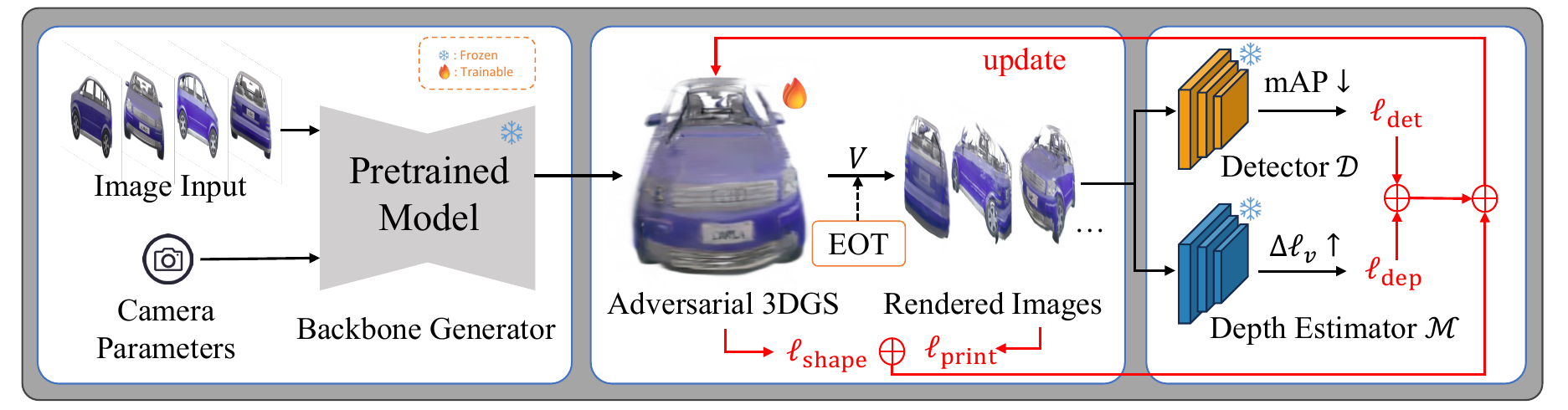}\vspace{-4mm}
    \caption{\textbf{BiTAA framework.} A frozen 3DGS backbone provides $\mathcal{G}_0$, and we optimize all 14 DoFs to obtain $\mathcal{G}$. With optional EOT, multi-view renderings feed (i) a detector, and (ii) a monocular depth head computing a signed log-depth bias within ROIs. Detection suppression, depth-bias, and lightweight regularizers are aggregated over views and back-propagated to update $\mathcal{G}$.}\vspace{-5mm}

    \label{fig:framework}
\end{figure*}

\subsection{Overview}
\label{sec:overview}

BiTAA performs a \emph{single} perturbation in the 3D Gaussian parameter space so that, after differentiable rendering from multiple viewpoints, the resulting images concurrently (i) suppress detection confidence and (ii) induce a signed, controllable bias in monocular depth within target regions. The 3DGS generation network is treated as a frozen backbone that provides an initial set $\mathcal{G}_0$; our optimization directly updates $\mathcal{G}$.

\textbf{Pipeline.}
As shown in Figure~\ref{fig:framework}, given a view set $V=\{v_1,\ldots,v_m\}$, we render $I_v=\mathcal{R}(\mathcal{G},v)$, optionally apply expectation-over-transformation (EOT) for physical reality on the detection branch, and feed the images to two frozen task models (detector and depth estimator). The overall objective couples a detection-suppression loss with a \emph{log-depth} bias loss restricted to a per-view ROI $\Omega_v$, along with regularizers enforcing shape stability and printability.

\subsection{Bi-Task Adversarial Framework}
\label{sec:framework}

\textbf{Forward-backward pipeline.}
Starting from the backbone output $\mathcal{G}_0$, we enable all 14 DoFs of the Gaussian set $\mathcal{G}$ and render each calibrated view $v\!\in\!V$:
\begin{equation}
I_v \;=\; \mathcal{R}(\mathcal{G},\,v), \qquad I^{0}_v \;=\; \mathcal{R}(\mathcal{G}_0,\,v).
\end{equation}
We then feed $I_v$ (and optionally its EOT-augmented version $\tau_t(I_v)$ with $t\!\sim\!\mathcal{T}$) to a frozen detector $\mathcal{D}$ and a frozen monocular depth estimator $\mathcal{M}$. The detector provides class-filtered boxes and scores for the detection branch; the depth estimator provides dense predictions $d_v=\mathcal{M}(I_v)$, while the clean depth $d^0_v=\mathcal{M}(I^0_v)$ is cached once per view for stable residuals. This forms a standard forward pass through the two tasks without training them.

\textbf{Per-view ROI for depth control.}
Depth manipulation is confined to semantically relevant regions on each view. Specifically, we construct a per-view ROI $\Omega_v$ from the detector's boxes after confidence filtering and an inward isotropic shrink (ratio $\rho\!\in\!(0,1)$) to suppress border noise. Equivalently, we use a binary mask $M_v$ with $\Omega_v=\{x\mid M_v(x)=1\}$. This ROI is only used to localize the depth-bias term; detection is still computed on the full image.

\textbf{View aggregation and optimization.}
Losses from the detection and depth branches are averaged over the view set $V$ (and over EOT samples if enabled) and combined with shape/printability regularizers; exact definitions are given in Sec.~\ref{sec:loss}. We update $\mathcal{G}$ by first-order optimization while keeping $\mathcal{R}$, $\mathcal{D}$ and $\mathcal{M}$ frozen:
\begin{equation}
\mathcal{G}\ \leftarrow\ \mathcal{G}\ -\ \eta\,\nabla_{\mathcal{G}}\,\mathcal{L}(\mathcal{G}).
\end{equation}
Two single-task variants are obtained by disabling one task term ($\lambda_{\mathrm{dep}}\!=\!0$ or $\lambda_{\mathrm{det}}\!=\!0$); they are used later to probe cross-task transfer, while the joint objective reveals interactions between detection and depth.

\subsection{Composite Loss Function}
\label{sec:loss}

BiTAA couples a detection-suppression loss with a signed log-depth bias (restricted to per-view ROIs), alongside regularizers for geometric stability and printability. All losses are aggregated over the view set $V$ and back-propagated to the 3D Gaussian parameters.

\textit{Detection suppression.}
For each view $v\!\in\!V$ and an EOT sample $t\!\sim\!\mathcal{T}$ with image operator $\tau_t$, let $p_v(t)$ denote the maximum confidence among vehicle-related classes returned by the frozen detector on the augmented image.
\begin{equation}
p_v(t) \;=\; \max_{k\in\mathcal{K}_v}\ s_{v,k}^{(t)} ,
\end{equation}
where $s_{v,k}^{(t)}$ is the confidence score of detection $k$ on $\tau_t(I_v)$ and $\mathcal{K}_v$ indexes retained detections after class/score filtering.
\begin{equation}
\mathcal{L}_{\mathrm{det}} \;=\; \frac{1}{|V|}\sum_{v\in V}\ \mathbb{E}_{t\sim\mathcal{T}}\!\left[-\log\!\big(1-p_v(t)+\delta\big)\right],
\end{equation}
with a small $\delta\!>\!0$ to avoid numerical issues. This form drives the strongest detection toward zero confidence.

\textit{Signed log-depth bias.}
Let $d_v$ and $d_v^{0}$ be the depth maps from the frozen estimator on $I_v$ and the clean rendering $I_v^{0}$, respectively (the latter cached once). Define the per-pixel log-depth residual $\Delta\ell_v(x)$ and average it within the per-view ROI $\Omega_v$ (skipping views with $|\Omega_v|=0$).
\begin{equation}
\Delta\ell_v(x) \;=\; \log\!\big(d_v(x)+\varepsilon\big)\ -\ \log\!\big(d_v^{0}(x)+\varepsilon\big),
\end{equation}
\begin{equation}
\mathcal{L}_{\mathrm{dep}} \;=\; \frac{1}{|V|}\sum_{v\in V}\ \frac{1}{|\Omega_v|}\sum_{x\in\Omega_v}\ \Big(\Delta\ell_v(x)\ -\ s\,\beta\Big)^{2},
\end{equation}
where $\varepsilon\!>\!0$ ensures stability, $s\!\in\!\{+1,-1\}$ controls the direction (push-far / pull-near), and $\beta\!>\!0$ sets the target bias magnitude.

\textit{Geometric stability.}
To preserve object reality, we penalize deviations of geometry-related parameters from the backbone output $\mathcal{G}_0=\{(\mu_i^0,\alpha_i^0,s_i^0,q_i^0,c_i^0)\}_{i=1}^{N}$. Let $\Delta\mu_i=\mu_i-\mu_i^0$, $\Delta s_i=s_i-s_i^0$, and $\Delta q_i=q_i-q_i^0$.
\begin{equation}
\begin{aligned}
\mathcal{L}_{\mathrm{shape}}
&= \sum_{i=1}^{N}\Big(
  w_\mu\|\Delta\mu_i\|_2^2
+ w_s\|\Delta s_i\|_2^2 
+ w_q\|\Delta q_i\|_2^2
\Big) \\
&\quad + \zeta\,\sum_{i=1}^{N}\big(\|q_i\|_2^2-1\big)^2 \,.
\end{aligned}
\end{equation}
where $\{w_\bullet\}$ weight per-parameter penalties and the last term softly enforces unit quaternions.

\textit{Printability.}
We constrain perturbations to remain modest in color-space and visually smooth in image-space. Let $\Delta c_i=c_i-c_i^0$ denote per-Gaussian color changes, and $R_v=I_v-I_v^{0}$ the residual image.
\begin{equation}
\mathcal{L}_{\infty} \;=\; \max\!\Big(0,\ \max_{i\in[1,N]}\ \max_{c\in\{r,g,b\}}\ |\Delta c_i^{(c)}|\ -\ \varepsilon_\infty\Big),
\end{equation}
\begin{equation}
\mathcal{L}_{\mathrm{TV}} \;=\; \frac{1}{|V|}\sum_{v\in V}\ \big\|\nabla R_v\big\|_{1},
\end{equation}
\begin{equation}
\mathcal{L}_{\mathrm{HF}} \;=\; \frac{1}{|V|}\sum_{v\in V}\ \big\langle W,\ |\mathcal{F}(R_v)|\big\rangle,
\end{equation}
where $\varepsilon_\infty$ is the color budget, $\nabla$ is the finite-difference gradient (isotropic TV), $\mathcal{F}$ is the DFT magnitude, and $W$ is a ring-shaped mask emphasizing high spatial frequencies. The realizability term combines these components:
\begin{equation}
\mathcal{L}_{\mathrm{print}} \;=\; \mathcal{L}_{\infty}\ +\ \alpha_{\mathrm{TV}}\,\mathcal{L}_{\mathrm{TV}}\ +\ \gamma_{\mathrm{HF}}\,\mathcal{L}_{\mathrm{HF}}.
\end{equation}

\textit{Overall objective.}
The full loss is a weighted sum of the foregoing components, aggregated over views (and EOT for detection):

\begin{equation}
\mathcal{L}_(\mathcal{G})=\lambda_{\mathrm{det}}\mathcal{L}_{\mathrm{det}}
+\lambda_{\mathrm{dep}}\mathcal{L}_{\mathrm{dep}}
+\lambda_{\mathrm{shape}}\mathcal{L}_{\mathrm{shape}}+ \lambda_{\mathrm{print}}\mathcal{L}_{\mathrm{print}}.
\end{equation}

We ignore the depth term on views with $|\Omega_v|=0$; EOT is applied only to the detection branch unless stated. All task networks remain frozen, and gradients flow through the renderer to $\mathcal{G}$.

\subsection{Cross-Task Transfer Protocol}
\label{sec:transfer}

To quantify how optimizing for one task transfers to the other, we define two single-task training protocols with bi-task evaluation, plus a joint baseline. All protocols use the same fixed view set $V$ and frozen task models.

\noindent\textbf{Det$\rightarrow$Depth.}
Train with detection-only loss by disabling the depth term:
\begin{equation}
\lambda_{\mathrm{det}}>0,\qquad \lambda_{\mathrm{dep}}=0.
\end{equation}
\emph{Evaluation.} Let $\mathrm{mAP}^{\mathrm{clean}}$ and $\mathrm{mAP}^{\mathrm{adv}}$ denote clean and adversarial mAP@0.5 aggregated over $V$. We report a task-normalized relative change (TNR) for detection,
\begin{equation}
\mathrm{TNR}_{\mathrm{det}}=\frac{\mathrm{mAP}^{\mathrm{clean}}-\mathrm{mAP}^{\mathrm{adv}}}{\mathrm{mAP}^{\mathrm{clean}}}\in[0,1],
\end{equation}
and the cross-task depth change using AbsRel (refer to \ref{sec:expsetup}),
\begin{equation}
\mathrm{TNR}_{\mathrm{depth}}=\frac{\mathrm{AbsRel}^{\mathrm{adv}}-\mathrm{AbsRel}^{\mathrm{clean}}}{\mathrm{AbsRel}^{\mathrm{clean}}}.
\end{equation}

\noindent\textbf{Depth$\rightarrow$Det.}
Train with depth-only loss by disabling the detection term:
\begin{equation}
\lambda_{\mathrm{dep}}>0,\qquad \lambda_{\mathrm{det}}=0.
\end{equation}
\emph{Evaluation.} We again compute $\mathrm{TNR}_{\mathrm{depth}}$ and the cross-task detection change $\mathrm{TNR}_{\mathrm{det}}$ as above.




\section{Experiment}
\label{sec:experiment}

\subsection{Experimental Setup}
\label{sec:expsetup}

\textbf{Victim models.} We evaluate bi-task transfer across three frozen detectors and three frozen monocular depth estimators under the same input resolution and preprocessing. \emph{Detectors} include Faster R-CNN~\cite{FasterRCNN}, Mask R-CNN~\cite{MaskRCNN}, and SSD~\cite{SSD}, all pretrained on COCO and run with standard score-thresholding. \emph{Depth estimators} include, Monodepth2~\cite{Monodepth2}, DPT-Large~\cite{DPT}, and Depth Anything V1 ~\cite{DepthAnything}. We use the authors' released weights and default inference resolutions. All victim models remain frozen throughout attack optimization.

\textbf{Baseline models.} We compare BiTAA with representative attacks on camera-based perception under a unified protocol. Detection baselines include CAMOU~\cite{CAMOU}, UPC~\cite{UPC}, DAS~\cite{DAS}, FCA~\cite{FCA}, and TT3D~\cite{TT3D}, covering digital and physical formulations commonly used for object detectors. Depth baselines include APA~\cite{APA}, SAAM~\cite{SAAM}, and 3D2FOOL~\cite{3D2FOOL}, which directly optimize depth-oriented objectives to distort estimated geometry. For fairness, all methods are run against the same frozen task models and input resolution, with author-recommended hyperparameters when available and minimal tuning on a held-out set; training steps and augmentation options are matched to our setting when applicable.

\textbf{Attack configuration.} We optimize the 3D Gaussian parameterization end-to-end, enabling all 14 degrees of freedom per Gaussian while keeping the renderer and task networks frozen. The composite objective uses fixed scalar weights to balance branches: $\lambda_{\mathrm{det}}$ and $\lambda_{\mathrm{dep}}$ for detection and depth, together with $\lambda_{\mathrm{shape}}$ and $\lambda_{\mathrm{print}}$ for regularization; unless otherwise specified, we set $(\lambda_{\mathrm{det}},\lambda_{\mathrm{dep}},\lambda_{\mathrm{shape}},\lambda_{\mathrm{print}})=(1.0,\,1.0,\,0.2,\,0.1)$. To control depth bias, we specify a signed target $(s,\beta)$, where $s\!\in\!\{+1,-1\}$ encodes the 'far' versus 'near' bias and $\beta\!\in\![0,0.10]$ sets the desired magnitude of the ROI-averaged log-depth shift; $\beta\!=\!0$ reduces to detection-only optimization, while larger $\beta$ increases the command strength of bias in a dose-responsive manner as validated in Sec.~\ref{sec:control}.

\textbf{Camera setup.} All renderings use square images at $512\times512$ with a shared pinhole intrinsic across views. For the physical evaluation, we use a $1{:}30$ Audi~A2 model and capture multi-view images with a Redmi K60 Pro smartphone camera at the scale-consistent distance: a real-world $5\,\mathrm{m}$ test range is emulated by placing the phone approximately $5/30\,\mathrm{m}\ (\approx16.7\,\mathrm{cm})$ from the model and sweeping azimuth/elevation to obtain varied viewpoints. The synthetic and physical settings therefore share aligned intrinsics (fixed $\mathrm{fov}_y$ and identical near/far) and controlled extrinsics, facilitating consistent cross-domain evaluation.

\textbf{Evaluation metrics.} For detection error, we report mAP@0.5--mean average precision at an IoU threshold of 0.5--averaged over the view set $V$. For depth estimation error, we use (i) Abs Relative difference (AbsRel), defined as $\frac{1}{|\Omega|}\sum_{x\in\Omega}\frac{|d(x)-d^{gt}(x)|}{d^{gt}(x)}$ ($\uparrow$), and (ii) RMSE (log), $\sqrt{\frac{1}{|\Omega|}\sum_{x\in\Omega}(d(x)-d^{gt}(x))^{2}}$ ($\uparrow$)~\cite{NIPS2014_91c56ce4,8100182}. In addition, to characterize the signed controllability central to our method, we measure the ROI-averaged log-depth displacement $\Delta\sigma$, computed per view as $\Delta\sigma_v=\frac{1}{|\Omega_v|}\sum_{x\in\Omega_v}\big[\log(d(x)+\varepsilon)-\log(d^{0}(x)+\varepsilon)\big]$ relative to the clean prediction $d^{0}$, and then aggregated across views. In subsequent comparisons with prior methods, we recommend using AbsRel as the primary depth metric (widely adopted and easy to interpret), including RMSE as a secondary indicator in the main table, and reserving $\Delta\sigma$ for our cross-task transfer, controllability, and sensitivity analyses where near/far bias and its magnitude are the focus.

\subsection{Baseline Comparison}

\begin{table}[t]
    \centering
    \caption{\textbf{Baseline Comparison} of BiTAA and existing baseline adversarial attack methods. Our method BiTAA achieves the best performance across all metrics, demonstrating superior adversarial effectiveness and depth estimation deviation.}\vspace{-2mm}
    \begin{tabular}{lccc}
        \toprule
        \multirow{2}{*}{Method} & \multicolumn{1}{c}{Det. metric} & \multicolumn{2}{c}{Depth metrics} \\
        \cmidrule(lr){2-2}\cmidrule(lr){3-4}
        & mAP (\%) $\downarrow$ & AbsRel $\uparrow$ & RMSE $\uparrow$ \\
        \midrule
        Vanilla                 & 74.68 & 0.1878 & 1.2906 \\
        \midrule
        CAMOU~\cite{CAMOU}      & 62.15 & 0.1881 & 1.3002 \\
        UPC~\cite{UPC}          & 64.82 & 0.1881 & 1.3010 \\
        DAS~\cite{DAS}          & 49.57 & 0.1874 & 1.2648 \\
        FCA~\cite{FCA}          & 28.24 & 0.1887 & 1.3022 \\
        TT3D~\cite{TT3D}        & 29.08 & 0.1893 & 1.3027 \\
        \midrule
        APA~\cite{APA}          & 58.36 & 0.1912 & 1.3092 \\
        SAAM~\cite{SAAM}        & 52.14 & 0.1926 & 1.3137 \\
        3D2Fool~\cite{3D2FOOL}     & 67.06 & 0.1952 & 1.3172 \\
        \midrule
            \textbf{BiTAA}          & \textbf{15.03} & \textbf{0.1989} & \textbf{1.3229} \\
        \bottomrule
    \end{tabular}\vspace{-3mm}
    \label{tab:benchmark}
\end{table}

Table~\ref{tab:benchmark} compares BiTAA with two families of baselines under a unified protocol (same frozen models, view set, and training budget). Detection-oriented attacks (CAMOU, UPC, DAS, FCA, TT3D) markedly reduce mAP@0.5 relative to the vanilla model (74.68\%), with FCA and TT3D reaching 28.24\% and 29.08\%, respectively, and an overall average of 46.77\% across the five methods (i.e., a mean drop of 27.91 points). However, their impact on off-task depth is small or inconsistent: the mean AbsRel over the five methods is 0.18832 (only +0.00052 vs.\ vanilla 0.1878), and the mean RMSE is 1.29418 (+0.00358 vs.\ 1.2906), with DAS even decreasing RMSE to 1.2648 (a -0.0258 change). This indicates that detector-only perturbations \textbf{transfer weakly} to depth and may occasionally regularize depth predictions. 

Depth-oriented attacks (APA, SAAM, 3D2FOOL) produce larger deviations on depth: AbsRel rises to 0.1912/0.1926/0.1952 (average +0.0052 over vanilla), and RMSE to 1.3092/1.3137/1.3172 (average +0.0228), but their effect on detection is moderate, with mAPs of 58.36\%, 52.14\%, and 67.06\% (average 59.19\%, a mean drop of 15.49 points). In contrast, BiTAA yields the strongest joint degradation: mAP falls to \textbf{6.05\%}, an absolute reduction of \textbf{68.63} points vs. vanilla and a further 22.19 points below the best detector-only baseline (FCA, 28.24\%); meanwhile AbsRel reaches \textbf{0.1989} (a +0.0111 increase, $\sim$2.1$\times$ the depth-only average gain) and RMSE \textbf{1.3229} (a +0.0323 increase, $\sim$1.4$\times$ the depth-only average gain). Taken together, the results substantiate our design goals: a single 3D perturbation that (i) surpasses detector-only attacks on their primary metric while simultaneously inducing (ii) substantially larger depth deviation than depth-only attacks, evidencing stronger cross-task coupling under the same budget and view setting.

\subsection{Controllability and Sensitivity}
\label{sec:control}

\begin{figure}
    \centering
    \includegraphics[width=1.0\linewidth]{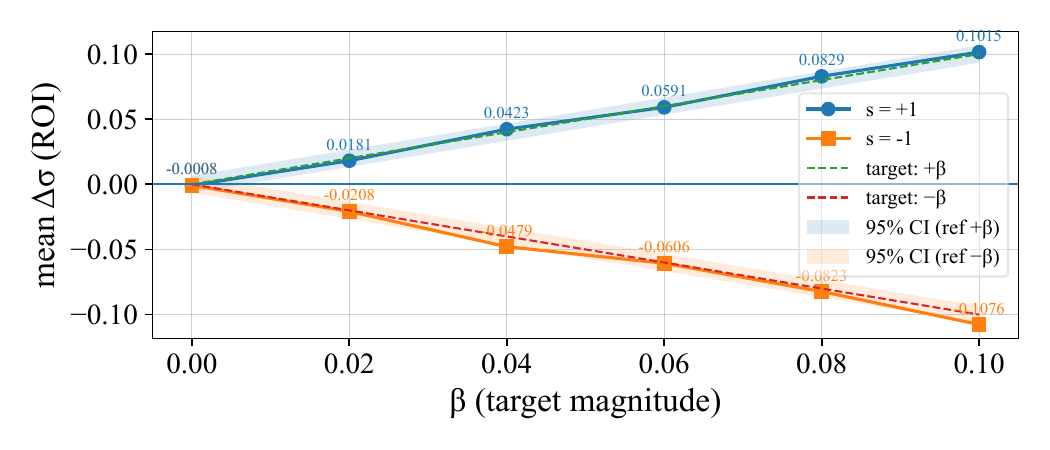}
    \vspace{-5mm}\caption{\textbf{Dose-response of signed log-depth bias.}}\vspace{-5mm}
    \label{fig:dose_response}
\end{figure}

\textbf{Signed log-depth bias is dose-controllable.} We evaluate the controllability promised by our composite loss (Sec.~\ref{sec:loss}) by sweeping the target magnitude $\beta\!\in\![0,0.10]$ under both bias directions $s\!\in\!\{+1,-1\}$ and reporting the ROI-averaged signed log-depth shift $\overline{\Delta\sigma}$ over multiple views/runs. Figure~\ref{fig:dose_response} visualizes the empirical means (scatter+line) together with the theoretical targets $\overline{\Delta\sigma}=s\cdot\beta$ (slope-$\pm1$ dashed lines) and their 95\% confidence bands estimated from residual scatter. The curves exhibit a clear, nearly linear dose-response: as $\beta$ increases, $\overline{\Delta\sigma}$ grows (or decreases) monotonically with slope close to one and negligible intercept, and the two signed branches ($s\!=\!+1$ for “far”, $s\!=\!-1$ for “near”) are approximately mirror-symmetric about the origin. The tight confidence bands indicate low variance across views and seeds, evidencing that the optimization remains stable despite multi-view aggregation and ROI masking.

\textbf{Implications for controllable misperception.} These results confirm Contribution~2: our bi-task formulation enables controllable, signed depth bias while jointly optimizing a detection-suppression objective. In practice, a target $s\!\cdot\!\beta$ translates to a multiplicative depth change $\exp(\overline{\Delta\sigma})$ within the ROI, allowing us to dial “near” or “far” misperception with predictable strength. We further observe that the calibration holds across representative models (e.g., DPT-L, DepthAnything, Monodepth2), with model-dependent sensitivity reflected only in the slope tightness rather than the trend itself. Overall, the dose-response behavior validates that a single perturbation in 3DGS space can produce depth shifts that (i) follow the commanded sign and magnitude, and (ii) do so consistently across views, providing a practical handle to probe safety margins in camera-only perception.

\begin{table}[t]
\centering
\caption{\textbf{Cross-task transfer: Det$\rightarrow$Depth.}}\vspace{-3mm}
\setlength{\tabcolsep}{6pt}
\begin{tabular}{lcccc}
\toprule
\multirow{2}{*}{Proxy Det} & \multicolumn{3}{c}{Target Depth (TNR$_\mathrm{depth}$, \%)} & \multirow{2}{*}{Row Mean} \\
\cmidrule(lr){2-4}
 & DPT-L & Mono2 & DA-V1 & \\
\midrule
Faster R-CNN & \textbf{22.82} & \textbf{9.06} & \textbf{9.66} & \textbf{13.85} \\
Mask R-CNN   & \textbf{31.44} & \textbf{19.06} & \textbf{33.74} & \textbf{28.08} \\
SSD          & \textbf{11.08} & \textbf{6.62} & \textbf{10.50} & \textbf{9.40} \\
\midrule
Col Mean     & \textbf{21.78} & \textbf{11.58} & \textbf{17.97} & \textbf{Overall 17.11} \\
\bottomrule
\end{tabular}
\label{tab:det2dep}
\end{table}
\begin{table}[t]
\centering
\caption{\textbf{Cross-task transfer: Depth$\rightarrow$Det.}}\vspace{-3mm}
\setlength{\tabcolsep}{6pt}
\begin{tabular}{lcccc}
\toprule
\multirow{2}{*}{Proxy Depth} & \multicolumn{3}{c}{Target Detector (TNR$_\mathrm{det}$, \%)} & \multirow{2}{*}{Row Mean} \\
\cmidrule(lr){2-4}
 & FR-CNN & MR-CNN & SSD & \\
\midrule
DPT-L        & \textbf{1.49} & \textbf{2.05} & \textbf{19.00} & \textbf{7.51} \\
Monodepth2   & \textbf{0.78} & \textbf{0.88} & \textbf{3.94} & \textbf{1.87} \\
DepthAnything & \textbf{7.44} & \textbf{5.94} & \textbf{8.86} & \textbf{7.41} \\
\midrule
Col Mean     & \textbf{3.24} & \textbf{2.96} & \textbf{10.60} & \textbf{Overall 5.60} \\
\bottomrule
\end{tabular}\vspace{-3mm}
\label{tab:dep2det}
\end{table}

\subsection{Cross-task Transferability} 
\label{sec:expcross}
\textbf{Cross-task transferability.} Under our unified protocol with single-task training and dual-task evaluation, we observe a clear directional asymmetry consistent with our motivation and abstract: attacks optimized for detection transfer strongly to depth, whereas depth-only attacks transfer more weakly to detection. Concretely, the macro-averaged depth degradation in the Det$\!\rightarrow\!$Depth grid reaches \(\overline{\mathrm{TNR}_{\mathrm{depth}}}=17.1\%\) across all models and views, while the macro-averaged detection degradation in the Depth$\!\rightarrow\!$Det grid is \(\overline{\mathrm{TNR}_{\mathrm{det}}}=5.6\%\). The ratio of these macro means is approximately \(3\times\), indicating substantially stronger transfer from detection to depth. These results align with our design of a bi-task framework and composite loss that expose cross-task couplings without changing the task networks, and they validate the claim in the introduction that the interaction between detection and depth is both measurable and asymmetric.

\textbf{Det$\!\rightarrow\!$Depth: which targets drive transfer?} When using detectors as proxies, two-stage models induce the largest cross-task effect on depth: Mask R-CNN attains the highest row mean (\(28.08\%\)), followed by Faster R-CNN (\(13.85\%\)) and SSD (\(9.40\%\)). On the target side, transformer-based DPT-L is most affected (\(21.78\%\) column mean), with DepthAnything V1 next (\(17.97\%\)) and Monodepth2 most resilient (\(11.58\%\)). We also note salient cells such as Mask R-CNN\(\!\rightarrow\!\)DepthAnything at \(33.74\%\). These patterns suggest that detector-driven perturbations alter object-centric cues (e.g., silhouette, shading, and local photometric consistency) that depth estimators--especially global-transformer backbones--heavily reuse, amplifying cross-task degradation. This observation directly supports our claim that a single 3D perturbation can bias monocular depth while suppressing detection under a standardized evaluation.

\textbf{Depth$\!\rightarrow\!$Det: where does transfer appear?} Using depth estimators as proxies yields smaller but non-negligible transfer to detection. DPT-L and DepthAnything V1 produce comparable row means (\(7.51\%\) and \(7.41\%\)), while Monodepth2 transfers the least (\(1.87\%\)), consistent with its weaker log-depth bias response in our ablations. As targets, SSD is notably more vulnerable (\(10.60\%\) column mean) than Faster/Mask R-CNN (\(3.24\%\) and \(2.96\%\)), and the cell DPT-L\(\!\rightarrow\!\)SSD peaks at \(19.00\%\). We attribute this to single-stage detectors' stronger reliance on dense texture and boundary evidence that is indirectly perturbed by depth-oriented optimization. Overall, while the average Depth$\!\rightarrow\!$Det transfer is smaller than Det$\!\rightarrow\!$Depth, the consistent, model-agnostic drops further underscore the practical risk of cross-task coupling in camera-only perception, motivating cross-task-aware defenses and evaluation protocols as advocated in our introduction.

\subsection{Ablation Study}

\subsubsection{Regularizers}
\begin{table}[t]
\centering
\caption{\textbf{Effect of Regularizers.}}\vspace{-3mm}
\setlength{\tabcolsep}{4.8pt}
\begin{tabular}{lcccc}
\toprule
Variant & TNR$_\mathrm{det}$ (\%) $\uparrow$ & mAP (\%) $\downarrow$ & LPIPS $\downarrow$ &  Side$\Delta$ ($\times 10^{3}$) $\downarrow$ \\
\midrule
Full (ours)            & 69.43 & 15.35 & \textbf{0.5396} & \textbf{1.7723} \\
$-\mathcal{L}_{\mathrm{print}}$ only  & 70.05 & 15.04 & 0.5436 & 1.7984 \\
$-\mathcal{L}_{\mathrm{shape}}$ only  & 71.80 & 14.16 & 0.5612 & 2.5537 \\
$-$ Both               & \textbf{73.16} & \textbf{13.48} & 0.5794 & 2.7921 \\
\bottomrule
\end{tabular}\vspace{-3mm}
\label{tab:ablation_regs}
\end{table}

\textbf{Regularizers trade adversarial strength for realism.} Table~\ref{tab:ablation_regs} quantifies the impact of the shape and printability terms in our composite loss. We deliberately report only detection-side metrics (TNR$_\mathrm{det}$ and mAP) together with perceptual/parametric realism because depth-side effects are not discriminative here: as established in Sec.~\ref{sec:control}, the signed log-depth bias reliably reaches the commanded target, making depth numbers less informative for ablation. Removing either regularizer yields a modest increase in attack strength (e.g., TNR$_\mathrm{det}$ $69.43\!\to\!71.80\%$ and mAP $15.35\!\to\!14.16$ for $-\mathcal{L}_{\mathrm{shape}}$), and dropping both pushes the strongest detection degradation (TNR$_\mathrm{det}$ $73.16\%$, mAP $13.48$). This confirms the intuitive trade-off: the terms do suppress a small portion of adversarial potency, but they are not the primary source of the effect.

\noindent\textbf{Perceptual and parametric stability justify the regularization.} In exchange for the small loss of raw attack strength, the regularizers preserve both image-level realism and 3DGS stability. LPIPS (referenced earlier) is lowest for the full model (0.5396) and increases as constraints are removed (0.5794 without both), indicating more noticeable artifacts outside ROIs. To directly measure how much the 3D asset is altered, we report \emph{Side$\Delta$}, which aggregates average parameter drift across the Gaussian set: Euclidean changes in position and rotation together with mean-squared changes in opacity, scale, and color, summarized as a single scalar ($\times 10^{-3}$; lower is better). Side$\Delta$ is smallest with all constraints (1.772), rises mildly without $\mathcal{L}_{\mathrm{print}}$ (1.798), and increases markedly without $\mathcal{L}_{\mathrm{shape}}$ (2.553) and without both (2.792), highlighting the role of the shape term in preventing geometry/appearance drift. Overall, the full setting achieves a favorable balance--strong detection degradation with noticeably better realism and parameter stability--so we retain both regularizers in all main results.

\subsubsection{EOT Modes.}

\begin{table}[t]
\centering
\caption{\textbf{EOT improves robustness under transformations.}}\vspace{-3mm}
\setlength{\tabcolsep}{5.5pt}
\begin{tabular}{lccccc}
\toprule
EOT & TNR$_\mathrm{det}$ $\uparrow$ & mAP (\%) $\downarrow$ & LPIPS $\downarrow$ & Var$_\mathrm{EOT}(\Delta\sigma)$ ($\times 10^{5}$) $\downarrow$ \\
\midrule
off   & 64.31 & 17.92 & 0.5837 & 1.5256 \\
partial & 68.04 & 16.05 & 0.5502 & 1.4830 \\
on    & \textbf{69.43} & \textbf{15.35} & \textbf{0.5396} & \textbf{1.4324} \\
\bottomrule
\end{tabular}\vspace{-3mm}
\label{tab:eot_modes}
\end{table}

\textbf{EOT strengthens transfer under test-time transformations.} Table~\ref{tab:eot_modes} evaluates attacks on the expected test distribution induced by random photometric/geometric transformations. Training with full EOT yields the strongest detection degradation, with the partial variant in between and off weakest (TNR$_\mathrm{det}$ $64.31\%$; mAP $17.92$). This ordering indicates that averaging gradients over transformations does not merely preserve digital strength--it can improve it when evaluation also involves realistic nuisances. In other words, EOT prevents overfitting to a canonical rendering and aligns the learned 3D perturbation with transformation-stable image evidence, which translates into higher transfer on the transformed test views.

\noindent\textbf{Stability and perceptual quality also benefit from EOT.} Beyond raw strength, EOT reduces the dispersion of the commanded log-depth shift across transformations: Var$_\mathrm{EOT}(\Delta\sigma)$ decreases from $1.5256\!\times\!10^{-5}$ (off) to $1.4324\!\times\!10^{-5}$ (on), indicating more consistent bias under the same ROI and target. At the image level, LPIPS monotonically improves (0.5837 $\rightarrow$ 0.5502 $\rightarrow$ 0.5396), suggesting that EOT encourages smoother, less artifact-prone appearance changes--consistent with the regularization effects observed in Sec.~\ref{tab:ablation_regs}. Taken together, these results support the use of EOT when the deployment environment introduces view and rendering variability: it offers stronger cross-task degradation on the relevant test distribution while simultaneously enhancing stability and perceptual realism.

\subsection{Physical Evaluation}
\label{sec:physical}

\begin{figure}[t]
\centering
\includegraphics[width=\linewidth]{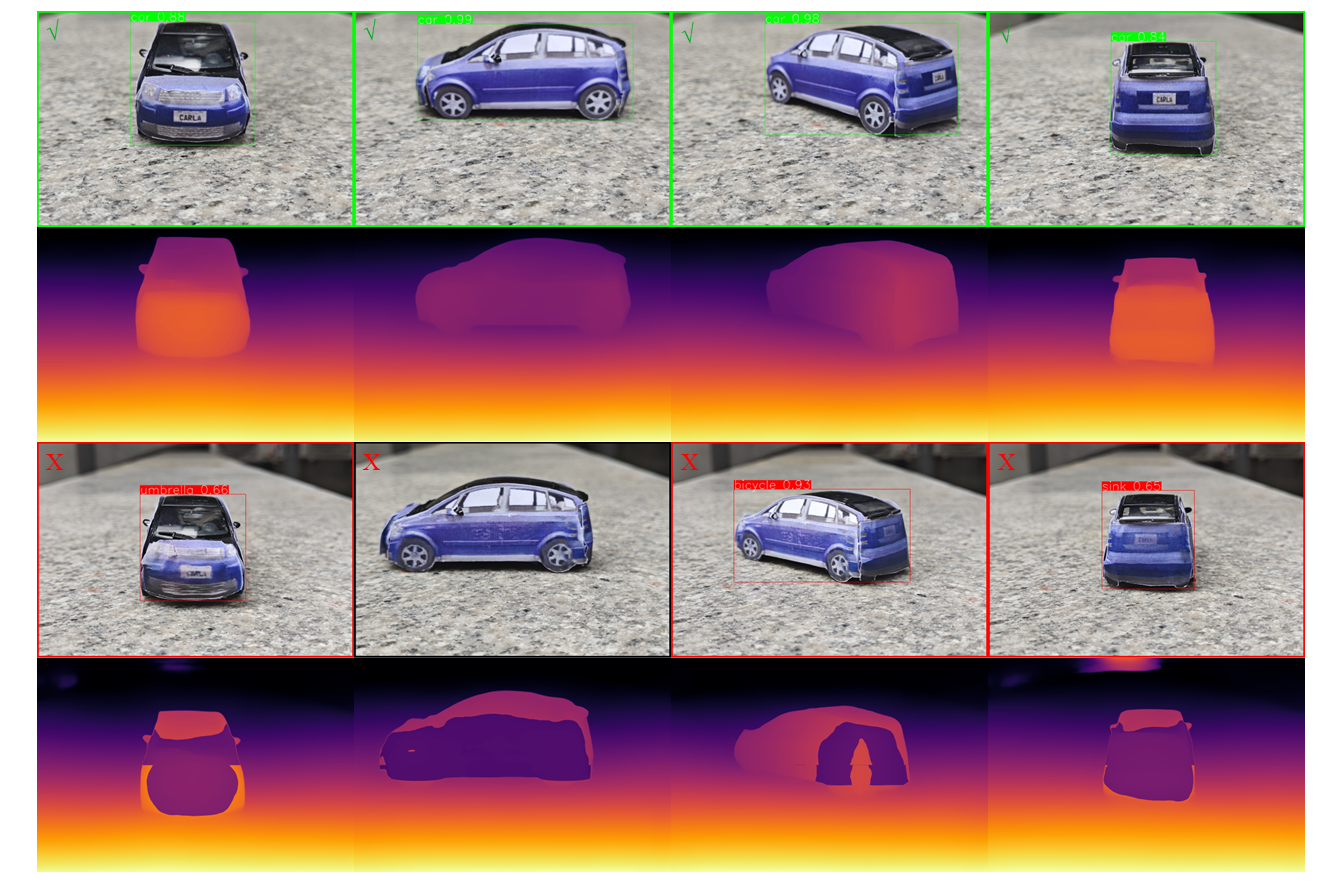}\vspace{-4mm}
\caption{\textbf{Real-world results at four viewpoints.} Columns (left$\rightarrow$right): clean image with detection overlays; clean depth heatmap; adversarial image with detection overlays; adversarial depth heatmap. Rows correspond to different azimuth viewpoints; the two depth maps in each row share the same colormap range.}\vspace{-6mm}
\label{fig:physical}
\end{figure}

\textbf{Setup and visualization.}
Figure~\ref{fig:physical} composes a $4{\times}4$ panel for four viewpoints of the 1:30 Audi~A2 model, each showing (from left to right) the clean image with detector outputs, its depth heatmap, the adversarial image with detector outputs, and the corresponding adversarial depth heatmap. We fix the detection thresholding/NMS across conditions and annotate the top class and confidence. For depth, the per-row heatmaps share an identical color scale so that clean$\leftrightarrow$adv comparisons are visually meaningful. Across all views, the clean images are confidently recognized as \textit{car} (e.g., scores ${\approx}0.98$), whereas the adversarial images display suppressed confidence or misclassification (e.g., \textit{umbrella}, \textit{bicycle}, \textit{sink}) under the same viewpoint, matching the intended detection degradation.

\textbf{Depth bias in the wild.}
The adversarial depth maps reveal signed, localized log-depth shifts within the visually salient regions of the vehicle body, while background regions remain largely unaffected. This regional effect is consistent with our training dynamics: although we did not manually constrain \emph{where} to modify, the 3DGS-based optimization preferentially updates texture-rich body panels (where the ROI has strong image gradients) and leaves the upper body (windows/roof) mostly unchanged. As a result, the induced bias manifests over the car body with the commanded sign, and persists across viewpoints despite lighting changes.

\textbf{Robustness of the pipeline.}
Notably, the physical textures used here were produced with a deliberately \emph{coarse} workflow (generic office printing, simple cutting/adhesion, minor misalignment and surface glare). Even under these non-ideal conditions, we observe consistent detection suppression and depth bias, indicating that the learned perturbation and our training protocol (including EOT) are tolerant to reasonable color reproduction errors, paper/gloss artifacts, and small pose/placement jitters. This aligns with our simulation findings that EOT trades a small amount of digital strength for markedly better stability under real nuisances.

\textbf{Path to deployment and limitations.}
A practical transfer path is straightforward: export the optimized perturbation as high-resolution texture segments and fabricate a matte \emph{adversarial wrap} that can be applied to target surfaces (with fiducial marks for registration, seam-aware tiling, and optional lamination). The minor visual imperfections visible in Fig.~\ref{fig:physical} (e.g., slight color banding or boundary artifacts) stem from the 3DGS backbone and consumer-grade printing rather than the attack itself; our framework is backbone-agnostic and can immediately benefit from future, higher-fidelity differentiable 3D reconstructions and better print pipelines. Overall, the results substantiate that a single 3D perturbation realized as a physical texture simultaneously degrades detection and imposes a controllable, signed depth bias in real scenes.


\section{Conclusion}
\label{sec:conclusion}
We presented \textbf{BiTAA}, a bi-task adversarial attack that operates directly in the 3D Gaussian Splatting (3DGS) parameter space and yields a \emph{single} perturbation capable of simultaneously suppressing detection and imposing a controllable, signed log-depth bias. Treating 3DGS as a frozen backbone, our dual-model framework couples a detector (with optional EOT for physical realism) and a depth estimator via a composite objective that integrates detection suppression with an ROI-restricted depth-bias term parameterized by $(s,\beta)$. This design enables \emph{controllability} (dose–response of the commanded bias) while maintaining \emph{stability} across views and tasks through lightweight shape/printability regularization. We further introduced a unified cross-task transfer protocol and task-normalized metrics, which consistently show cross-task degradation and a clear \emph{asymmetry} between Det$\!\rightarrow$Depth and Depth$\!\rightarrow$Det transfer. Real-world experiments with printed textures on a scaled model corroborate the digital findings, demonstrating robustness to common physical nuisances.

Looking ahead, our results motivate \emph{cross-task-aware} defenses for camera-centric autonomy and open avenues to (i) stronger, higher-fidelity differentiable 3D backbones, (ii) broader task sets and multi-sensor perception, (iii) adaptive ROI generation and safety-aware constraints, and (iv) principled, standardized protocols for evaluating multi-task transfer under realistic transformations. We release code and protocols to facilitate reproducible research and to encourage community progress on both attack and defense.



\section*{ACKNOWLEDGMENT}
Specific funding information will be provided in the camera-ready version.

{
    \footnotesize
    \bibliographystyle{IEEEtran}

    \bibliography{main}
}

\end{document}